\documentclass[conference]{IEEEtran}
\IEEEoverridecommandlockouts

\usepackage{cite}
\usepackage{amsmath} 
\usepackage{amsfonts, dsfont} 
\usepackage{bbold}
\usepackage{hyperref} 
\usepackage{graphicx} 
\usepackage{booktabs,xcolor}
\usepackage[ruled,vlined]{algorithm2e}
\graphicspath{ {./images/} }
\def\BibTeX{{\rm B\kern-.05em{\sc i\kern-.025em b}\kern-.08em
T\kern-.1667em\lower.7ex\hbox{E}\kern-.125emX}}

\title{Last layer state space model for representation learning and uncertainty quantification}

\author{\IEEEauthorblockN{Max Cohen}
	\IEEEauthorblockA{\textit{T\'el\'ecom SudParis, CITI, TIPIC} \\
		Institut Polyechnique de Paris \\
		maxjcohen@proton.me}
	\and
	\IEEEauthorblockN{Maurice Charbit}
	\IEEEauthorblockA{\textit{Accenta, Boulogne-Billancourt} \\
		maurice.charbit@accenta.fr}
	\and
	\IEEEauthorblockN{Sylvain Le Corff}
	\IEEEauthorblockA{\textit{T\'el\'ecom SudParis, CITI, TIPIC} \\
		Institut Polyechnique de Paris}
}

\begin{document}
\maketitle
\begin{abstract}
	As sequential neural architectures become deeper and more complex, uncertainty estimation is more and more challenging.
	Efforts in quantifying uncertainty often rely on specific training procedures, and bear additional computational costs due to the dimensionality of such models.
	In this paper, we propose to decompose a classification or regression task in two steps: a representation learning stage to learn low-dimensional states, and a state space model for uncertainty estimation.
	This approach allows to separate representation learning and design of generative models.
	We demonstrate how predictive distributions can be estimated on top of an existing and trained neural network, by adding a state space-based last layer whose parameters are estimated with Sequential Monte Carlo methods.
	We apply our proposed methodology to the hourly estimation of Electricity Transformer Oil temperature, a publicly benchmarked dataset.
	Our model accounts for the noisy data structure, due to unknown or unavailable variables, and is able to provide confidence intervals on predictions.
\end{abstract}


\section{Introduction}
\label{sec:intro}

Recurrent Neural Networks (RNN) were first introduced as an efficient and convenient architecture to address short time dependencies problems.
They have been consistently improved to develop longer term memory, and optimize their implementations \cite{Bengio1994LearningLD,Hochreiter1997LongSM}. 
Current deep learning frameworks allow stacking arbitrary high number of recurrent layers, whose parameters are estimated by gradient descent through automated differentiation procedures, as shown in \cite{Graves2013SpeechRecognition}.
However, many critical applications, such as medical diagnosis or drug design discovery, require not only accurate predictions, but a good estimate of their uncertainty (\cite{Crowson2016AssessingCalibration, Mervin2020UncertaintyQuantification}).
Fostering the dissemination of deep learning-based algorithms to such fields requires to design new approaches for uncertainty quantification.

Bayesian statistics are able to approximate the distributions of future observations and to provide uncertainty estimation \cite{Hinton1995BayesianLF}.
Several architectures inspired by Variational Inference (VI, see \cite{Jordan2004AnIT}) emerged by considering latent states as random variables and approximating their posterior distribution.
The authors of \cite{10.5555/3157096.3157343} built on a traditional recurrent architecture by modelling temporal dependencies between these latent random states.
Results presented in \cite{Fortunato2017bayesian} yield improved performances when considering local gradient information for computing the posterior.
In \cite{Cohen2023VariationalDiscreteLatent}, a prior model based on a Markov chain is estimated in the latent space of an Auto Encoder in order to compute uncertainty estimation on the observation.

Sequential Monte Carlo (SMC) methods have also been successfully applied to Recurrent Neural Networks.
Instead of computing a single latent vector at each time step, a set of particles representing the distribution of the latent space are propagated, and associated with importance weights.
In \cite{maddison2017filtering,naesseth2017variational}, the authors were able to model complex distributions on dependant data.
We turn to \cite{Martin2020TheMC} for an example using more complex neural architectures, such as the Transformer.

In \cite{Blundell2015}, the authors considered weights as random variables and proposed approximations of their posterior distributions allowing more robust predictions. Such Bayesian neural networks have been proposed and studied in a variety of works, see for instance \cite{hernandez2015probabilistic,teye2018bayesian}. However, these methods are computationally intensive for high dimensional models and we do not have statistical guarantees on their ability to capture the target  posterior distribution, see \cite{NEURIPS2020_b6dfd418}.

Monte Carlo Dropout (MC Dropout) methods offer to capture uncertainty by leveraging Dropout during both training and evaluation tasks, producing variable predictions from a single trained recurrent model, see \cite{Gal2016NIPS}.
In the recent years, MC Dropout methods have been applied in many industrial fields, such as flight delay prediction \cite{Vandal2018} or molecular simulations \cite{Wen2020UncertaintyQI}.
Alternatively, ensemble methods consist in training distinct networks to obtain a combined prediction, as shown in \cite{Pearce2018}.
However, these frequentist approaches fail to guarantee proper calibration of the model, as highlighted by \cite{ashukha2020pitfalls}, and suffer various limitations, see \cite{Fong2020}.

In an effort to provide an alternative strategy with limited computation overhead, \cite{Brosse2020OnLA} suggests splitting representation learning and uncertainty estimation to solve classification problems for independent data.
A deep classifier is first trained to obtain task dependent representations of the data, on which ensemble models are fitted to approximate the distribution of the observations.
Their experiments indicate that performing uncertainty estimation on the last-layer of the model outperforms baseline networks and is an appealing trade-off between computational cost and uncertainty quantification.
However, as this method is restricted to independent data, it cannot be directly applied to time series.

Inspired by \cite{Brosse2020OnLA}, we propose a last layer approach to split uncertainty quantification from representation learning, in the context of dependent data.
This new method combines high expressivity, quality uncertainty estimations and ease of training.
Our main contributions are as follows.
\begin{itemize}
	\item We propose a decoupled architecture composed of an arbitrary sequential model and a state space model layer.
	\item This last layer allows to introduce complex predictive distributions for the observations.
	      Its parameters are estimated through approximate samplings using Sequential Monte Carlo methods, as the likelihood of the observations is not available explicitly in such a setting.
	\item Our methodology allows for arbitrary deep architectures, and does not suffer the overconfidence of frequentist approaches.
\end{itemize}

\section{Last layer decoupling}
\begin{figure}[htpb]
	\centering
	\includegraphics[width=0.5\linewidth]{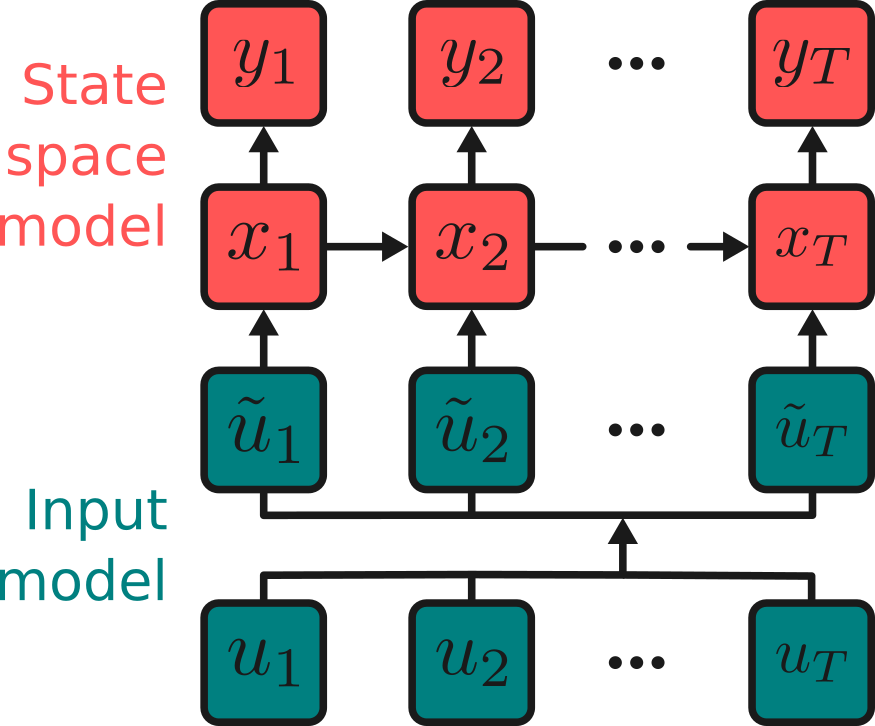}
	\caption{Our architecture combining a generic input model with a state space model on the last layer.}
	\label{fig:architecture}
\end{figure}

Estimating the parameters of potentially high-dimensional models with unobserved (i.e. noisy) layers is a challenging task.
We therefore propose to first train an input model following traditional deep learning approaches, then use Monte Carlo methods in a lower dimensional state space to account for uncertainty, with tractable and computationally efficient simulation-based methods.
The two-stage training algorithm is presented in Algorithm~\ref{alg:particle_filter}, and the architecture of the model is described in Figure~\ref{fig:architecture}

In the following, for any sequence $(a_m,\ldots, a_n)$ with $n\geq m$, we use the short-hand notation $a_{m:n} = (a_m,\ldots, a_n)$.
Let $T\ge 1$ be a given time horizon.
We consider a regression task with observations $Y_{1:T}$ associated with inputs $U_{1:T}$.

\subsection{Representation learning}%
In this paper, we consider an arbitrary multi-layer neural network $h_\varphi$ with unknown parameters $\varphi$, responsible for extracting high level features from the input time series:
\begin{align*}
	\widetilde U_{1:T} & = h_\varphi(U_{1:T})\,, & \text{input model.}
\end{align*}
We produce an estimate $\hat \varphi$ during the first training stage, by introducing an auxiliary function $\kappa_\psi$ to model the observations as follows: for all $1 \leq k \leq T$, $Y_k = \kappa_\psi(Y_{k-1}, \widetilde U_k) + \epsilon_k$ and $Y_0 = \kappa_\psi(\widetilde U_0) + \epsilon_0$, where $(\epsilon_k)_{k\leq 0}$ are independent centered Gaussian random variables with unknown variance.
The input model is trained on a simple deterministic regression task, by performing gradient descent on the mean squared error, leading to a first estimate of $\varphi$ and $\psi$.
We keep the estimated parameters $\hat \varphi$ while the auxiliary function $\kappa_\psi$ and its parameters, only designed to model the observations, are discarded.

\subsection{State space model}
\label{sub:proposed_architecture}
The next step is to define  a state space model taking as input the previously extracted features $\widetilde U_{1:T}$.
Let $X_{1:T}$ a sequence of stochastic hidden states computed recursively and $Y_t$ their associated predictions.
For all $k \geq 1$, the model is defined as:
\begin{align*}
	X_k & = g_\theta(X_{k-1}, \widetilde U_k) + \eta_k\,, & \text{state model, }       \\
	Y_k & = f_\theta(X_k) + \epsilon_k\,,                 & \text{observation model, }
\end{align*}
where $\theta$ are the unknown real-valued parameters of the network (weights and biases) and $f_\theta$ and $g_\theta$ are nonlinear parametric functions.
We chose $(\eta_k)_{k\geq 1}$ and $(\epsilon_k)_{k\geq 1}$ as two sequences of independent centered Gaussian random variables with covariance matrices $\Sigma_x$ and $\Sigma_y$, although any distribution can be substituted.

This decoupled approach aims at reducing the number of parameters in $\theta$, compared to $\varphi$, in order to estimate them using Sequential Monte Carlo methods.
In the next section, we describe this second training procedure for the last layer only, by keeping $\hat \varphi$ fixed.

\section{Sequential Monte Carlo Layer}%
\label{sub:uncertainty_estimation}
In this section, we detail how to estimate the parameters $\theta$, $\Sigma_x$ and $\Sigma_y$ in the model introduced in Section~\ref{sub:proposed_architecture}, from a record of observations $Y_{1:T}$.
This is challenging because the likelihood of the observations is not available explicitly, as it would require integrating over the hidden states $X_{1:T}$.
Consequently, the score function is intractable.
We propose to optimize a Monte Carlo estimator of this score function, using Fisher's identity \cite{monographie-randal}:
\begin{equation}
	\nabla_\theta \log p_\theta(Y_{1:T}) = \mathbb{E}_\theta \left[ \nabla_\theta\log p_\theta(X_{1:T}, Y_{1:T}) | Y_{1:T} \right]\,,
	\label{eq:grad_ll}
\end{equation}
where $\mathbb{E}_\theta$ designs the expectation under the model parameterized by $\theta$ (the dependency on the input $U_{1:T}$ is kept implicit here for better clarity).
In the following paragraphs, we denote by $\Psi_{\mu, \Sigma}$ the Gaussian probability density function with mean vector $\mu$ and covariance matrix $\Sigma$.

\subsection{Particle filter}
The conditional distribution of $X_{1:T}$ given $Y_{1:T}$ is not available explicitly for a nonlinear state space model, but it can be approximated using a family of $N$ particles $(\xi^{\ell}_{1:T})_{\ell=1}^N$ associated with importance weights $(\omega^{\ell}_T)_{\ell=1}^N$.
At $k = 0$, $(\xi^{\ell}_0)_{\ell=1}^N$ are sampled independently from $\rho_0 = \Psi_{0, \Sigma_x}$, and each particle $\xi^{\ell}_0$ is associated with the standard importance sampling weight $\omega_0^{\ell} \propto  \Psi_{Y_0, \Sigma_y}(f_\theta(\xi^{\ell}_0))$.
Then, for $k\geq 1$, using $\{(\xi^{\ell}_{k-1},\omega^{\ell}_{k-1})\}_{\ell=1}^N$, we sample pairs $\{(I^{\ell}_k,\xi^{\ell}_{k})\}_{\ell=1}^N$ of indices and particles from the instrumental distribution:
\[
	\pi_{k}(\ell,x) \propto \omega_{k-1}^{\ell} p_k(\xi^{\ell}_{k-1},\widetilde U_k,x)\,.
\]
In this application we use for $p_k(\xi^{\ell}_{k-1},\widetilde U_k,\cdot)$ the prior kernel $\Psi_{g_\theta(\xi^\ell_{k-1}, \tilde U_k), \Sigma_x}$.
For $\ell \in \{1,\ldots,N\}$, $\xi^{\ell}_k$ is associated with the importance weight $\omega^{\ell}_k \propto \Psi_{Y_k, \Sigma_y}(f_\theta(\xi^{\ell}_k))$. Such a particle filter with multinomial resampling is referred to as the bootstrap algorithm, see \cite{gordon1993novel}. It has been extended and analyzed in many directions in the past decades, see \cite{pitt1999filtering,douc2005comparison,Chopin_2020}. In other lines of works, the adaptive tuning of the Monte Carlo effort has been analyzed in order to  adapt the
number of particles on-the-fly, see \cite{elvira2016adapting,elvira2021performance}.

\subsection{Particle smoother and online estimation}
Our framework allows the use of any particle smoother to estimate \eqref{eq:grad_ll}.
In this paper, we first describe the Path-space smoother \cite{Kitagawa1996} for its simplicity, in order to illustrate our approach.
In practice, it often leads to particle path degeneracy \cite{Andrieu2005}, which can be mitigated by substituting a more complex smoother such as the Forward Filtering Backward Smoothing \cite{Doucet2000OnSM} or the Forward Filtering Backward Simulation algorithm \cite{Godsill2004MonteCS}.
Additionally, because estimating \eqref{eq:grad_ll} amounts to computing a smoothed expectation of an additive functional, we can also use very efficient forward-only SMC smoothers such as the PaRIS algorithm and its pseudo-marginal extensions \cite{Olsson2014EfficientPO,gloaguen2022pseudo}.
With $\xi^i_{1:T}$ the ancestral line of $\xi^i_{T}$, the score function \eqref{eq:grad_ll} can be estimated as follows using automated differentiation:
$$\widehat {S}^N_\theta(Y_{1:T}) = \sum_{\ell=1}^N \omega_T^\ell\nabla_\theta\log p_\theta(\xi^\ell_{1:T}, Y_{1:T})\,,$$
where $p_\theta$ is the joint probability density function of $(X_{1:T}, Y_{1:T})$ for the model described in Section~\ref{sub:proposed_architecture}.

The degeneracy relative to the smoothing problem can be overcome using
\emph{backward sampling}. It is specifically designed for additive functionals so it is well suited to our setting \eqref{eq:grad_ll} since
$\nabla_\theta\log p_\theta(x_{1:T}, y_{1:T}) = \sum_{t=1}^T \nabla_\theta \log m_{\theta}(x_{t-1},\widetilde u_t;x_t)  r_{\theta}(x_{t},y_t)$,
where $m_{\theta}(x_{t-1},\widetilde u_t;\cdot)$ is the transition density of the state model and $r_{\theta}(x_{t},\cdot)$ is the density of the conditional distribution of $y_t$ given $x_t$ and by convention $m_{\theta}(x_{0},\widetilde u_1;\cdot) = \rho_0(\cdot)$.

The Monte Carlo estimator of the score function can be obtained online  by setting,
$	\widehat {S}^N_\theta(y_{1:T}) = \sum_{i = 1}^N \omega_T^i\tau_T^i$,
where the statistics $\{\tau_s^i\}_{i = 1}^N$ satisfy the recursion $\tau_{s + 1}^{i} = \tau_{s}^{I_{s+1}^{i}} + \tilde{h}_s(\xi_{s}^{I_{s+1}^{i}}, \xi_{s + 1}^{i})$,
where $\tilde{h}_s(x_s,x_{s+1}) = \nabla_\theta \log m_{\theta}(x_{s},\widetilde u_{s+1};x_{s+1}) r_{\theta}(x_{s+1},y_{s+1})$.
Following \cite{DelMoral2010,Olsson2014EfficientPO,martin2020backward,gloaguen2022pseudo}, the degeneracy of the path-space smoother can be overcome by performing an online {\it PaRis} update of the statistics $\tau_{s + 1}^{i}$, $1\leq i\leq N$, using the backward kernel of the hidden Markov chain.

An appealing application of the last layer approach is recursive maximum likelihood estimation, i.e., where new observations are used only once to update the estimator of the unknown parameter $\theta$. In \cite{Brosse2020OnLA}, the authors used in particular Stochastic Gradient Descent (SGD) and Stochastic Gradient Langevin Dynamics to update the estimation of $\theta$ and perform uncertainty quantification. In state space models, recursive maximum likelihood estimation produces a sequence $\lbrace\theta_k\rbrace_{k\geq 0}$ of parameter estimates writing, for each new observation $Y_{k},~k\geq 1$,
$$
	\theta_{k} = \theta_{k-1} + \gamma_k \nabla_\theta \ell_{\theta}(Y_k | Y_{0:k - 1}) \,,
$$
where $\ell_{\theta}(Y_k | Y_{0:k - 1})$ is the loglikelihood for the new observation given all the past, and $\lbrace\gamma_k\rbrace_{k\geq 1}$ are positive step sizes such that $\sum_{k \geq 1}\gamma_k = \infty$ and $\sum_{k \geq 1}\gamma_k^2 < \infty$. The practical implementation of such an algorithm, where $\nabla_\theta\ell_{\theta}(Y_k | Y_{0:k - 1})$ is approximated using the weighted samples $\{(\xi^{\ell}_k,\omega^{\ell}_k)\}_{\ell=1}^N$ can be found for instance in \cite{gloaguen2022pseudo}. The PaRIS algorithm proposed in \cite{Olsson2014EfficientPO} allows to use the weighted samples $\{(\xi^{\ell}_k,\omega^{\ell}_k)\}_{\ell=1}^N$ and the statistics $\{\tau^{\ell}_k\}_{\ell=1}^N$ on-the-fly to approximate $\nabla_\theta \ell_{\theta}(Y_k | Y_{0:k - 1})$.

Although this algorithm is very efficient to update parameters recusrively, it is computationally intensive and therefore fits particularly well our last layer approach as it would be intractable for very high dimensional latent states.

\begin{algorithm}
	\label{alg:particle_filter}
	$\hat \varphi \gets$ Train the input model $h_\varphi$\;
	$\widetilde U_{1:T} \gets h_{\hat \varphi}(U_{1:T})$\;
	Initialize parameter estimate $\widehat \theta_0$\;
	\For{$p \gets 1$ \KwTo $\mathrm{MaxIt}$}{
	$\xi_0^\ell \sim \rho_0$ and $\omega_0^{\ell} \propto  \Psi_{Y_0, \Sigma_y}(f_{\widehat \theta_{p-1}}(\xi^{\ell}_0))$\;
	\For{$k \gets 1$ \KwTo $T$}{
	\For{$j \gets 1$ \KwTo $N$}{
		$I_k^j \sim \mathbb{P}(I_k^j=m) = \omega_{k-1}^m$\;
		$\xi_k^j \sim p_k(\xi_{k-1}^{I_k^j},  \widetilde U_k,\cdot)$\;
		$\omega_k^j \propto  \Psi_{Y_k, \Sigma_y}(f_{\widehat \theta_{p-1}}(\xi^{j}_k))$\;
		Set $\xi_{0:k}^j = (\xi_{0:k-1}^{I_k^j},\xi_k^j)$.
		}
	}
	Update the parameter estimate using gradient descent with estimated gradient $\widehat {S}^N_{\widehat \theta_{p-1}}(Y_{1:T})$.
	}
	\caption{Two-stage learning}
\end{algorithm}

\section{Experiments}
\label{sec:exp}
\subsection{Data and model}
\label{sub:data}
We benchmarked our approach on the public Electricity Transformer Temperature (ETT) Dataset, designed in \cite{Zhou2021Informer} to forecast Oil temperature based on hourly power load records (ETTh1 subset).
%

\textbf{The Input model} is a $L=3$ layered GRU model, as defined in the deep learning framework PyTorch\footnote{\href{https://pytorch.org/docs/stable/generated/torch.nn.GRU.html}{https://pytorch.org/docs/stable/generated/torch.nn.GRU.html}}: for all $1 \leq \ell \leq L$ and all $1 \leq k \leq T$,
\begin{align*}
	r^\ell_k        & = \sigma(W_{ir} U^{\ell - 1}_k + b_{ir} + W_{hr} U^{\ell}_{k-1} + b_{hr}) \,,                \\
	z^\ell_k        & = \sigma(W_{iz} U^{\ell - 1}_k + b_{iz} + W_{hz} U^{\ell}_{k-1} + b_{hz}) \,,                \\
	n^\ell_k        & = \mathrm{tanh}(W_{in} U^{\ell - 1}_k + b_{in} + r^\ell_k (W_{hn} U^\ell_{k-1} + b_{hn}))\,, \\
	\tilde U^\ell_k & = (1-z^\ell_k) n^\ell_k+z^\ell_k U^\ell_{k-1}\,,
\end{align*}
where $\varphi = \{(W_{is}, b_{is}, W_{hs}, b_{hs}), s \in \{r, z, n\}\}$ are unknown parameters, and $\sigma: x \mapsto 1/(1+\mathrm{e}^{-x})$ is the sigmoid function.
The first layer of the network is assimilated to the input vectors, $\widetilde U_t^0 \equiv U_k$ and $\widetilde U^\ell_0 \equiv 0$.
The input dimension $d_\textnormal{in}=6$ corresponds to the number of power load records of the dataset, we set the output dimension to 6.
In order to estimate the parameters $\varphi$, we introduce an auxiliary GRU layer responsible for computing oil temperature predictions.
During the training, we minimize the cost function $\mathcal{L}_{\mathrm{input}}(\varphi) = \sum_{i=1}^{N_{\texttt{sample}}} \|\texttt{model}_{\varphi}(U^i_{1:T}) - Y^i_{1:T}\|^2$
between each sample of the dataset and the associated prediction obtain with this deterministic model.

\textbf{The State Space model} is implemented using PyTorch implementations of  RNN and Linear layers.
We chose the following form for $f_\theta$ and $g_\theta$:
\begin{align*}
	g_\theta & : X_{k-1}, \widetilde U_k \mapsto \tanh(W_{gx} X_{k-1} + b_{gx} + W_{gu} \widetilde U_k + b_{gu})\,, \\
	f_\theta & : X_k \mapsto  \sigma(W_f X_k + b_f)\,,
\end{align*}
where $\theta = \{W_{gx}, b_{gx}, W_{gu}, b_{gu}, W_f, b_f,\Sigma_x,\Sigma_y\}$ are unknown parameters.
All following experiments are conducted with $N=100$ particles and a batch size of 32, using the Adam optimizer introduced in \cite{Kingma2015AdamAM}.
The learning rate was chosen using a simple grid search.
We train models for a maximum of 50 epochs, and employ early stopping to prevent overfit.

\subsection{Evaluations}%
\label{sub:evaluations}

In this section, we illustrate the ability of our model to capture the distribution of future observations, by evaluating the benchmarked models using the following protocol. We draw 48 hours long samples $(u_{1:48}, y_{1:48})$ from the validation dataset, composed of a 24 hour long lookback window $(u_{1:24}, y_{1:24})$, containing historic commands and observations, and a predictions window where only future commands are available $(u_{25:48})$.
Each model produces $N=100$ 24 hour long forecasts $(y_{25:48}^{(i)})_{i=1}^N$.
We compute the Root Mean Squared Error (RMSE) between the observations and the average of the forecasts: $\mathrm{RMSE}^2 = T^{-1} \sum_{k=1}^T (Y_k - N^{-1} \sum_{i=1}^N y_k^{(i)})^2$.
Additionally, we evaluate the Prediction Interval Coverage Probability (PICP, see \cite{Durga2006PICP}) which measures the ratio of observations falling between a 95\% confidence interval: $\mathrm{PICP} = T^{-1} \sum^{T}_{k=1} \mathbb{1}_{[y_L^k, y_U^k]}(Y_k)$, where $y_U^k$ (resp. $y_L^k$) is the upper (resp. lower) bound of the confidence intervals.
Both criteria are reported in Table~\ref{tab:ci_comparison}.

\begin{figure}[htpb]
	\centering
	\includegraphics[width=\linewidth]{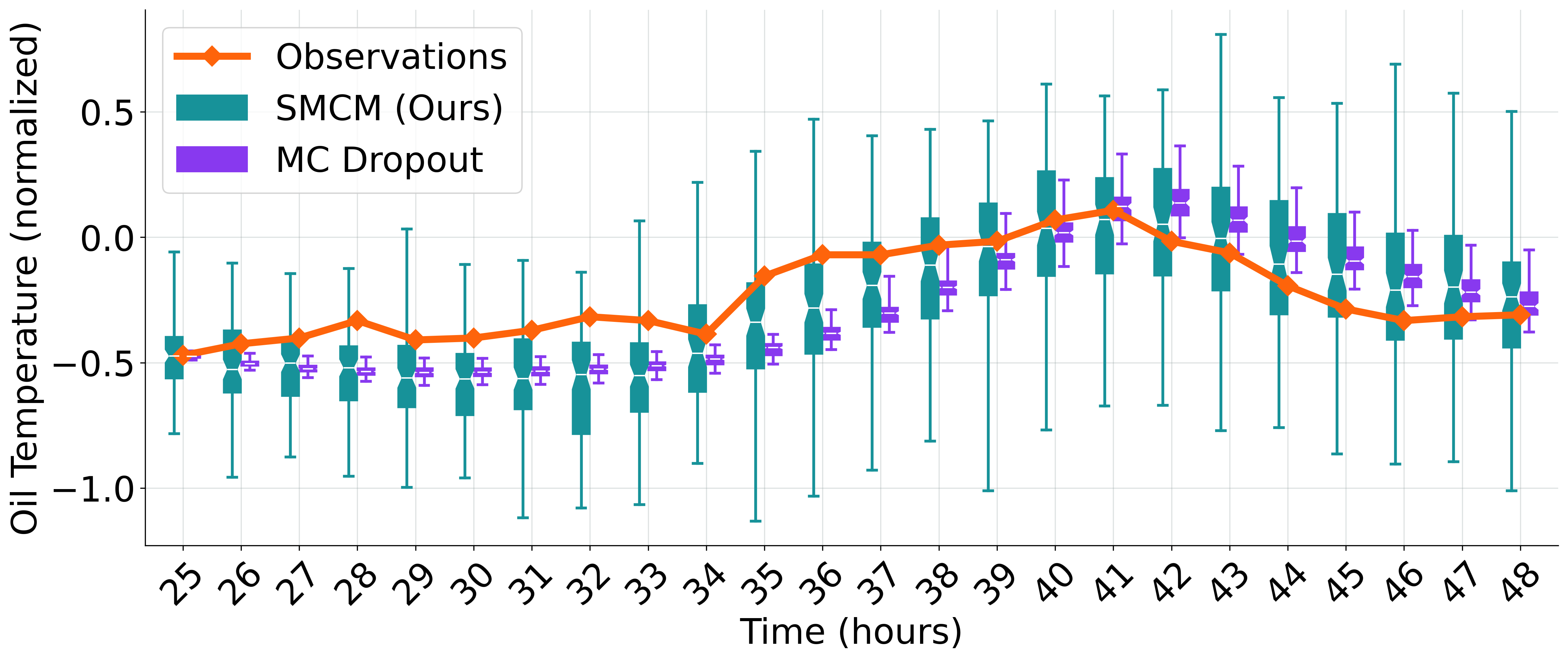}
	\caption{Forecasting of oil temperature given observations in the lookback window ($k<24$).
		Since resampling of particles is no longer available at that point, the uncertainty grows for our model.
		As a comparison, we plotted the confidence intervals produced by the MC Dropout model (for $p_\textnormal{drop}=0.01$).}
	\label{fig:filter_k+24}
\end{figure}

For our proposed model, predictions can be performed by approximating the predictive density $p_{\theta,\varphi}(y_{k+1}|U_{1:k+1},Y_{1:k})$ by
$$
	p^N_{\widehat\theta,\widehat\varphi}(y_{k+1})= \sum_{i=1}^{N}\omega_k^i p_{\widehat\theta,\widehat\varphi}(y_{k+1}|\xi_k^i,U_{k+1})\,,
$$
where $ p_{\widehat\theta,\widehat\varphi}(y_{k+1}|\xi_k^i,U_{k+1})$ is the predictive distribution of $Y_{k+1}$ described in Section~\ref{sub:proposed_architecture}.
In order to explore longer ranges, we run our model to get $N$ samples for any time horizon.
The associated intervals containing 95\% of the samples are displayed in Figure~\ref{fig:filter_k+24}, for 24 hours forecasts.

We compared our model with MC Dropout methods, by implementing recurrent dropout layers as described in \cite{Gal2016NIPS}.
The optimal dropout rate $p_\textnormal{drop}=0.01$ that we tuned by grid search is smaller than the proposed value in the original paper, which may be due to our much longer time series, similarly to results presented in \cite{Zhu2017DeepAC}.
Additionally, we evaluate the model with $p_\textnormal{drop}=0.05$, which we show slightly degrades performances.
The training procedure is similar to traditional recurrent models ; during inference, we draw $100$ samples from the dropout layers, and compute the same average forecasts and  intervals as for our model.
Despite being based on the same deep learning architecture, the MC dropout model is still largely overconfident, while our proposed model provide more credible empirical confidence intervals.
We also experimented with a Gaussian linear Hidden Markov Model (HMM) whose parameters are estimated with the Kalman smoother using the Expectation Maximization (EM, \cite{Dempster77EM}) algorithm.
Out of a range of possible latent dimension sizes $d_{\texttt{hmm}} \in \{1, 2, 4, 6\}$, we selected $d_{\texttt{hmm}}=4$ as it yielded the best performances.

\begin{table}[htpb]
	\centering
	\caption{Comparison of RMSE, PICP and computation time of our model against the benchmarked MC Dropout methods and HMM.
		Two versions of the dropout model were evaluated, with dropout values $p_\textnormal{drop}=0.05$ and $p_\textnormal{drop}=0.01$.
		Mean values of the estimators are displayed along with their variance.}
	\begin{tabular}{llll}
		\toprule
		             & RMSE            & PICP   & Computation time  \\
		\toprule
		SMCL (ours)  & $0.24 \pm 0.13$ & $98\%$ & $210 ms \pm 75.7$ \\
		MCD $p=0.01$ & $0.25 \pm 0.15$ & $59\%$ & $193 ms \pm 60.4$ \\
		MCD $p=0.05$ & $0.28 \pm 0.15$ & $65\%$ & $193 ms \pm 60.4$ \\
		HMM          & $0.44 \pm 0.13$ & $85\%$ & $994 ms \pm 21.4$ \\
		\bottomrule
	\end{tabular}
	\label{tab:ci_comparison}
\end{table}


\section{Conclusion}%
\label{sec:conclusion}

In this paper, we introduced a decoupled architecture for uncertainty estimation on a time series dataset.
Our deep neural network backbone is responsible for extracting high level features, while particle filtering in the last layer allows modelling recurrent nonlinear uncertainty.
Our proposed model does not suffer from the overconfidence of MC Dropout methods, while significantly improving on the performances of Hidden Markov Models.

We demonstrate the potential behind implementing latent space models as a modified RNN cell ;
more complex architectures, such as the GRU network used in the input model, or LSTM cells, could be considered.

Our decoupled architecture enables incorporating uncertainty estimation to an already trained network.
This opens the door to multiple, cheap finetuning of last layers parameters, from a global pretraining.

\clearpage
\bibliographystyle{IEEEtran}
\bibliography{refs.bib}
\end{document}